\pdfoutput=1

\documentclass[11pt]{article}

\usepackage[]{acl}

\usepackage{graphicx}
\usepackage{times}
\usepackage{latexsym}
\usepackage{multirow}
\usepackage{array}
\usepackage{graphicx}
\usepackage{graphicx}
\usepackage{float}
\usepackage{hyperref}

\usepackage[T1]{fontenc}

\usepackage[utf8]{inputenc}

\usepackage{microtype}

\usepackage{inconsolata}

\title{\textbf{L3Cube-IndicNews: News-based Short Text and Long Document Classification Datasets in Indic Languages}}

\author{Aishwarya Mirashi$^{1,3}$,  Srushti Sonavane$^{1,3}$, Purva Lingayat$^{1,3}$, Tejas Padhiyar$^{1,3}$ \and Raviraj Joshi$^{2,3}$ \\
        Pune Institute of Computer Technology, Pune$^1$ \\ Indian Institute of Technology Madras, Chennai$^2$ \\ L3Cube Labs, Pune$^3$}

\begin{document}

\maketitle
\begin{abstract}
In this work, we introduce L3Cube-IndicNews, a multilingual text classification corpus aimed at curating a high-quality dataset for Indian regional languages, with a specific focus on news headlines and articles. We have centered our work on 11 prominent Indic languages, including Hindi, Bengali, Marathi, Telugu, Tamil, Gujarati, Kannada, Odia, Malayalam, Punjabi and English. Each of these news datasets comprises 10 or more classes of news articles. L3Cube-IndicNews offers 3 distinct datasets tailored to handle different document lengths that are classified as: Short Headlines Classification (SHC) dataset containing the news headline and news category, Long Document Classification (LDC) dataset containing the whole news article and the news category, and Long Paragraph Classification (LPC) containing sub-articles of the news and the news category. We maintain consistent labeling across all 3 datasets for in-depth length-based analysis. We evaluate each of these Indic language datasets using 4 different models including monolingual BERT, multilingual Indic Sentence BERT (IndicSBERT), and IndicBERT. This research contributes significantly to expanding the pool of available text classification datasets and also makes it possible to develop topic classification models for Indian regional languages. This also serves as an excellent resource for cross-lingual analysis owing to the high overlap of labels among languages. The datasets and models are shared publicly at \url{https://github.com/l3cube-pune/indic-nlp}.

Keywords: Low Resource Languages, Indic Languages, Web Scraping, News Article Datasets, BERT, Short Text Classification, Long Documents. 
\end{abstract}

\section{Introduction}

India boasts a rich linguistic diversity, with over 700 languages spoken, out of which 22 are officially recognized. Some of the primarily spoken languages include Hindi, Bengali, Marathi, Telugu, Tamil, Gujarati, Kannada, Odia, Malayalam, Punjabi and English. Despite their widespread use, there's a notable scarcity of comprehensive Indic language datasets, primarily due to their low-resource status and linguistic complexity \cite{patil2017mars}.

These Indic languages are widely spoken all over India and have abundant data available on news websites, and social media but it’s difficult to find a labeled dataset of news headlines, articles, and their categories for text classification. This disparity has hindered progress in machine learning and Natural Language Processing (NLP) research for these languages.

Most Indic languages share similarities, but they utilize different writing scripts, making it more challenging to accurately predict news categories. Since English is used widely around the world, and many researchers have studied how to classify text in English, it has resulted in a surplus of classification datasets. The same is not the case for Indic languages. Although there are existing datasets for Indian languages, they have some limitations, such as a lesser number of categories or inconsistent classification criteria for news articles, which narrows down the scope of research in this field. While the IndicNLP News Article dataset covers the major languages it is limited by the count of target labels and high accuracy \cite{kakwani2020indicnlpsuite,kulkarni2022experimental}. This calls for the creation of more intricate datasets to effectively assess the performance of models. In essence, there is a need for complex datasets that can thoroughly evaluate model effectiveness. 

Text classification, a critical task in both machine learning and natural language processing (NLP), involves categorizing text documents into predefined classes based on their content. While there are a handful of publicly available datasets related to news, they often lack diversity in categories and sources, potentially leading to biased results.

Transformer-based models, like LongFormer\footnote{\url{https://huggingface.co/docs/transformers/model\_doc/longformer}} \cite{beltagy2020longformer}, require datasets with varying sequence lengths due to their sensitivity to text length. This highlights the need for specialized datasets to develop models for these low-resource languages. Hence, we introduce L3Cube-IndicNews, a comprehensive IndicNews Classification Dataset, sourced from diverse news websites specifically targeting low-resource languages.

This dataset encompasses over 3 lakh records, distributed across 12 diverse news categories, offering an extensive resource for supervised text classification. Each language dataset contains more than 26,000 rows, covering at least 10 significant news categories.

L3Cube-IndicNLP\footnote{\url{https://github.com/l3cube-pune/indic-nlp}} presents monolingual and multilingual models tailored to each Indian regional language. The repository includes individual BERT models for the languages focused on in this work. We conduct a comparative analysis of various monolingual and multilingual BERT models, including L3Cube monolingual BERT \cite{joshi2022l3cube_hind,joshi2022l3cube}, monolingual SBERT \cite{deode2023l3cube,joshi2023l3cube_mahasbert}, IndicSBERT (multilingual) \cite{deode2023l3cube}  and IndicBERT\cite{kakwani2020indicnlpsuite}.

The key contributions of this work are as follows:
\begin{itemize}
    \item Introduction of L3Cube-IndicNews, an extensive document classification dataset spanning ten significant Indian languages, each consistent with a range of 12 target labels. The dataset can also be used for news article headline-generation tasks.
    \item The corpus comprises three sub-datasets (IndicNews- SHC, LPC, and LDC) catering to short, medium, and long documents, each with varying sentence lengths but consistent target labels.
    \item The datasets are bench-marked using state-of-the-art pre-trained BERT models: L3Cube monolingual BERT, L3Cube monolingual SBERT, L3Cube IndicSBERT (multilingual) and IndicBERT. The models for individual languages are shared publicly on Hugging Face\footnote{\url{https://huggingface.co/l3cube-pune/marathi-topic-all-doc-v2}}(see Appendix).
\end{itemize}

\section{\textbf{Related work} }

IndicNLP News Article Classification\footnote{ https://github.com/AI4Bharat/indicnlp\_corpus\#indicnlp-news-article-classification-dataset } dataset is part of the AI4Bharat-IndicNLP Dataset \cite{kunchukuttan2020ai4bharat} that consists of news articles in 10 Indian languages categorized into classes like sports, entertainment, business, politics, and lifestyle. While this dataset contains a substantial number of records, it falls short in terms of the variety of categories available for news articles in each language, thereby limiting its diversity.

Multi Indic Languages News Dataset\footnote{ {https://www.kaggle.com/datasets/shaz13/multi-indic-languages-news-dataset}{https://www.kaggle.com/datasets/shaz13/multi-indic-languages-news-dataset}} is a dataset publicly available on Kaggle. It is a multi-language news dataset from Times Internet for various Indian languages. This data contains columns named title, link, description, long\_description, id. Despite the extensive size and diversity of this dataset, encompassing a rich collection of records across numerous Indian languages, it lacks language-wise and news category-wise segregation, which hampers clarity and ease of use.

Varta\footnote{\textsuperscript{https://huggingface.co/datasets/rahular/varta} } is a multilingual dataset for headline generation. It encompasses 41.8 million news articles in 14 Indic languages and English. This data is sourced from DailyHunt. This dataset is well-organized and includes a large number of records for each language, covering all the minor details. However, this dataset contains only those articles written by DailyHunt’s partner publishers resulting in a biased nature towards a particular narrative or ideology that can affect the representativeness and diversity of the dataset. \cite{aralikatte2023v} From every language, they randomly sample 10,000 articles each for validation and testing. On average, Varta articles have 17 sentences, and the headlines have just over one sentence. A typical article sentence contains about 18 words, and a headline sentence contains 11 words. While the dataset is a large-scale, high-quality dataset for Indic languages, the headlines in this dataset are 39\% smaller than the average sentence in an article. 

iNLTK\footnote{ \textsuperscript{https://github.com/goru001/inltk} } is an openly accessible dataset primarily comprising data for 13 Indic languages, sourced from Wikipedia articles. It encompasses over 12,000 cleaned rows for each of these languages \cite{arora2020inltk}. This dataset comprises publicly accessible data for languages like Hindi, Bengali, Punjabi, Kannada, and Oriya. For languages such as Gujarati, Malayalam, Marathi, and Tamil, they have created their dataset by extracting information from Wikipedia articles. While the dataset boasts a substantial volume of records for each language, it falls short in terms of categorizing the data into specific news categories. Furthermore, in some languages, it includes not only news-related articles but also other types of content, leading to inefficiency and inconsistency in the dataset's content and structure.  

ACTSA\cite{mukku2017actsa} focuses on building a gold-standard annotated corpus of Telugu sentences to support Telugu Sentiment Analysis. The raw data is scraped from five different Telugu news websites viz. Andhrabhoomi, Andhrajyothi, Eenadu, Kridajyothi and Sakshi. In total, they collected over 453 news articles which were then filtered down to 321 articles relevant to their work.

\begin{table} [H]
\centering
 \resizebox{78mm}{50mm}{%
\begin{tabular}{|c|c|p{7.5cm}|p{1.2cm}|}
\hline
\textbf{Language} & \textbf{Datasets available} & \textbf{Categories present}& \textbf{Articles}\\
\hline

 Hindi& BBC Articles& India, International, Entertainment,  Sports, Others& 172K \\
& BBC Hindi News Articles & India, Pakistan, News, International, Sntertainment, 
port, Science, China, Learningenglish, Social, Southasia, Business, Institutional, Multimedia& 4335 \\
\hline

 Bengali& AI4Bharat-IndicNLP Dataset &  Entertainment, Sports& 14K \\
& Soham Articles& Kolkata, State, National, International, Sports, Entertainment& 72K \\
\hline

Marathi& AI4Bharat-IndicNLP Dataset & Entertainment, Lifestyle, Sports& 4.5K \\
& iNLTK Headlines &State, Entertainment, Sports& 85K \\
\hline

 Telugu& AI4Bharat-IndicNLP Dataset & Entertainment, Business, Sports& 24K \\
\hline

Tamil& AI4Bharat-IndicNLP Dataset & Entertainment, Politics, Sports& 11.7K \\
& iNLTK Headlines& Tamil-cinema, Business, Spirituality & 127K \\
\hline

 Gujarati& AI4Bharat-IndicNLP Dataset & Business, Entertainment, Sports& 2K \\
& iNLTK Headlines & Entertainment, Business, Tech & 31K \\
\hline

 Kannada& AI4Bharat-IndicNLP Dataset & Entertainment, Lifestyle, Sports& 30K \\
& iNLTK-IndicNLP News Category& Entertainment, Sports, Tech& 6.3K \\
 \hline
 
Odia& AI4Bharat-IndicNLP Dataset & Business, Crime, Entertainment, Sports & 30K \\
& iNLTK-IndicNLP News Category& Sports, Business, Entertainment& 19K \\
\hline

 Malayalam& AI4Bharat-IndicNLP Dataset & Business, Entertainment, Sports, Technology & 6K \\
& iNLTK Headlines&Entertainment, Sports, Business  & 12K \\
\hline

 Punjabi& AI4Bharat-IndicNLP Dataset & Business, Crime, Entertainment, sports & 3.1K \\
&
 iNLTK -IndicNLP News Category & Politics, Non-politics & 800 \\
\hline

\end{tabular}
}
\caption{\label{citation-guide}
 Available Datasets in Indic languages 
}
\end{table}

\section{\textbf{Curating the dataset}}

We introduce L3Cube-IndicNews, a comprehensive dataset compilation designed to facilitate the classification of both short text and long documents. Within IndicNews, there are three meticulously crafted supervised datasets: Short Headlines Classification (SHC), Long Document Classification (LDC), and Long Paragraph Classification (LPC).

Featuring a broad spectrum of information, the dataset comprises a minimum of 10 distinct categories corresponding to 10 prominent languages spoken in India. The careful organization of both language and newly defined categories enhances overall clarity. Sourced from multiple websites, it ensures a diverse array of content, exclusively from reputable news sources.  Notably, there are no constraints on the length of articles and headlines, emphasizing a commitment to quality. In essence, it stands as a high-quality, versatile, and meticulously curated dataset. 

\subsection{Data collection}

For the dataset curation process, we identified several websites to collect a substantial number of articles for each news category. The Hindi language dataset was scraped from Jansatta\footnote{ https://www.jansatta.com/}. The Marathi language dataset was scraped from Lokmat\footnote{ https://www.lokmat.com/}. The Bengali language dataset was scraped from Aajkal\footnote{ https://www.aajkaal.in/}, Ganashakti\footnote{ https://ganashakti.com/}, BBC\footnote{ https://www.bbc.com/bengali}, Anandabazar\footnote{ https://www.anandabazar.com/}, abnews24\footnote{ https://www.sangbadpratidin.in/}, and Sangbadpratidin\footnote{ https://www.abnews24.com/}. The Telugu language dataset was scraped from ABP Telugu\footnote{ https://telugu.abplive.com/}. The Tamil Language dataset was scraped from Hindu Tamil\footnote{ https://www.hindutamil.in/}. The Gujarati language dataset was scraped from ABP Gujarati\footnote{ https://gujarati.abplive.com/}. The Kannada language dataset was scraped from Kannada Prabha\footnote{ https://www.kannadaprabha.com/} and PublicTV\footnote{ https://publictv.in/}. The Odia language dataset was scraped from Odisha Bhaskar\footnote{ https://odishabhaskar.com/} and  Dharitri\footnote{ {https://www.dharitri.com/}{https://www.dharitri.com/} }. The Malayalam language dataset was scraped from Madhyamam\footnote{ https://www.madhyamam.com/}. The Punjabi language dataset was scraped from Khabarwaale\footnote{ https://www.khabarwaale.com/}.  

The data was gathered through the utilization of the urllib package for managing URL requests, coupled with the BeautifulSoup package for extracting data from the HTML content of the requested URL. Every website had organized its news articles into predefined categories, and during the scraping process, we retained this categorization to utilize it as the target label.

Each dataset was originally scraped to contain 3 columns: Title, Category, and News. Then 35\% of the news article from column ‘News’ was used to create a fourth column ‘Sub article’, primarily containing a subset of news articles. The final curated dataset underwent shuffling and cleaning. 
We then divided these datasets into three supervised datasets: Short Headlines Classification (SHC), Long Document Classification (LDC), and Long Paragraph Classification (LPC).

\textit{\textbf{Short Headlines Classification (SHC):}} This dataset contains the headlines of news articles paired with their respective categorical labels.

\textit{\textbf{Long Paragraph Classification (LPC)}}\textbf{:} In this dataset, each record contains a sub-article of news with its respective categorical label.

\textit{\textbf{Long Document Classification (LDC)}}\textbf{:} This dataset contains records having an entire news article with its corresponding categorical label.

\subsection{Data Preprocessing} 

At first, we break down the article into sentences, making sure we handle the punctuation and sentence boundaries in Indic languages correctly. Then, we carefully break each sentence into tokens and filter them for special characters. We selectively retain tokens with an initial character aligned with the specific Indic language character set. We also created a dataset-cleansing function wherein each text is tokenized and then undesired elements, including words with specific characters, substrings, or minimal length are eliminated. Furthermore, regular expression was used to filter out words that don't match the script of the language. This data refinement process ensures the dataset is free from unwanted elements, making it optimal for further use.  
 
\subsection{Dataset Statistics}

Each dataset consists of more than 26,000 rows and 10-12 categories of news articles per dataset. Each of the 10 datasets was split into the train, test, and validation datasets in the ratio of 80:10:10.

 \begin{table}[H]
 \centering
 \resizebox{70mm}{48mm}{%

\begin{tabular}{|c|p{5cm}|} \hline  
          \textbf{Language} & \textbf{Labels}\\ \hline  
          Hindi (11)& Auto, Business, Crime, Education, Entertainment, Health, International, Nation, Politics, Sports, Technology\\ \hline  

          Bengali (10)& Sports, National, International, Kolkata, State, Politics,  Entertainment, Technology, Editorial, Lifestyle\\ \hline
          
          Marathi (12)& Auto, Bhakti, Crime, Education, Fashion, Health, International, Manoranjan, Politics, Sports, Tech, Travel \\ \hline

          Telugu (10)& Business, Crime, Education, Entertainment, Jobs, Lifestyle, Politics, Sports, Technology, World\\ \hline  
          
          Tamil (10)& Auto, Business, Crime, Education, Entertainment, Health, India, Lifestyle, Politics, World\\ \hline  

          Gujarati (10)& Astro, Auto, Business, Crime, Entertainment, International, Nation, Sports, State, Technology\\ \hline  
                  
          Kannada (10)& Business, Crime, Cuisine, Entertainment, International, Nation, Politics, Sports, State, Technology\\ \hline           
          
          Odia (10)& Lifestyle, Entertainment, News, Crime, Business, Health, Politics, Career, Sports, Editorial\\ \hline  
          
          Malayalam (10)& Crime, Entertainment, Gulf, International, Kerala, Lifestyle, National, Opinion, Sports, Technology\\ \hline                   
          Punjabi (10)& World, Sports, Religious, Education, Transfer \& Appointments, Literature, Health, National, Crime, Lifestyle\\ \hline         
          English (10)& Health, Business,  Elections, Education, Lifestyle, World, Sports, Entertainment, Science, Auto\\ \hline
     \end{tabular}
     }
     \caption{Languages and their categorical labels}
     \label{tab:my_label}
 \end{table}

\begin{table}[H]
    \centering
    \resizebox{0.5\textwidth}{!}{%
    \begin{tabular}{ccccc}
    \hline
         \textbf{Language}&  \textbf{Train}&  \textbf{Test}&  \textbf{Validation}& \textbf{Total}\\
         \hline
         Hindi&  30851&  3835&  3835& 38521\\
         Bengali&  23970 &  2997 &  2996 & 29963 \\
         Marathi&  22014&  2761&  2750& 27525\\
         Telugu&  21103&  2640&  2650& 26393\\
         Tamil&  25030&  3129&  3129& 31288\\
         Gujarati&  26472&  3347&  3417& 33236\\
         Kannada&  24642&  3058&  3058& 30758\\
         Odia&  27420&  3445&  3434& 34299\\
         Malayalam&  28000&  3500&  3500& 35000\\
         Punjabi&  25494&  3189&  3186& 31869\\
         English&  36877&  4610&  4610& 46097\\
        \hline
    \end{tabular}
    }
    \caption{Distribution of dataset into train, test, and validation in the ratio 80:10:10.}
    \label{tab:my_label}
\end{table}

\begin{figure}[H]
    \centering
    \includegraphics[width=8.0cm, height=1.2cm]{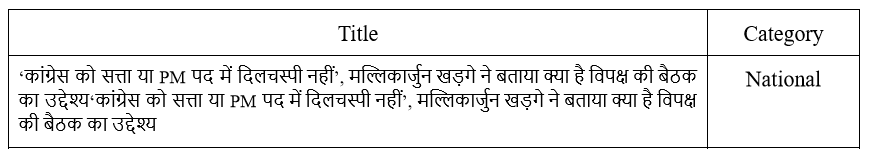}
    \caption{SHC Dataset Overview}
    \label{fig:enter-label}
\end{figure}

\begin{figure}[H]
    \centering
    \includegraphics[width=8.0cm]{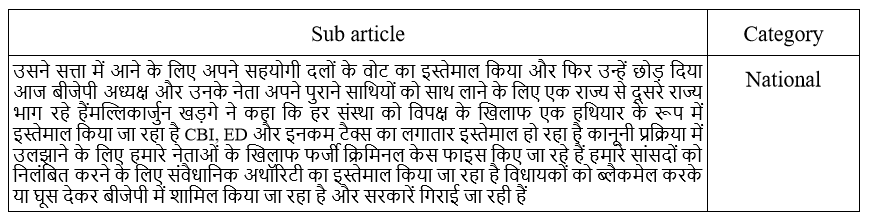}
    \caption{LPC Dataset Overview}
    \label{fig:enter-label}
\end{figure}
\begin{figure}[H]
    \centering
    \includegraphics[width=8.0cm, height=5.0cm]{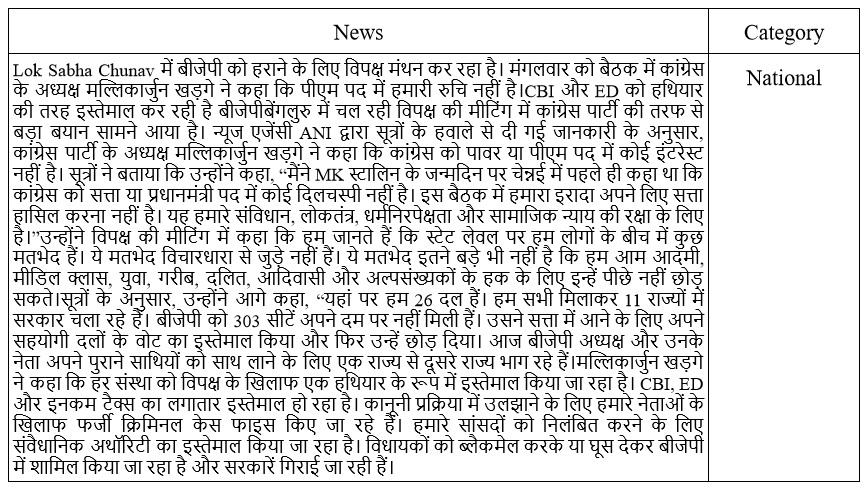}
    \caption{LDC Dataset Overview}
    \label{fig:enter-label}
\end{figure}

\section{Models}

\subsection[L3Cube Monolingual BERT]{L3Cube Monolingual BERT\footnote{ https://arxiv.org/abs/2211.11418} for Indic languages}

We use the monolingual BERT models for the 10 Indic languages, released by L3cube-Pune\footnote{ https://github.com/l3cube-pune} as the base models. These models are termed as HindBERT\footnote{ https://huggingface.co/l3cube-pune/hindi-bert-v2}, BengaliBERT\footnote{ https://huggingface.co/l3cube-pune/bengali-bert}, MahaBERT\footnote{ https://huggingface.co/l3cube-pune/marathi-bert}, TeluguBERT\footnote{ https://huggingface.co/l3cube-pune/telugu-bert}, TamilBERT\footnote{ https://huggingface.co/l3cube-pune/tamil-bert}, GujaratiBERT\footnote{ https://huggingface.co/l3cube-pune/gujarati-bert}, KannadaBERT\footnote{ https://huggingface.co/l3cube-pune/kannada-bert}, OdiaBERT\footnote{ https://huggingface.co/l3cube-pune/odia-bert}, MalayalamBERT\footnote{ https://huggingface.co/l3cube-pune/malayalam-bert}, PunjabiBERT\footnote{ https://huggingface.co/l3cube-pune/punjabi-bert}. These models are fine-tuned on the existing multilingual models like MuRIL \cite{khanuja2021muril}, xlmRoBERTa \cite{conneau2019unsupervised}, and IndicBERT on the monolingual corpus. \newline
\newline

\subsection[L3Cube Indic Sentence BERT models]{L3Cube Indic Sentence BERT\footnote{https://arxiv.org/pdf/2304.11434.pdf} models (Monolingual)}

We also evaluate L3Cube monolingual Indic SBERT models that are HindSBERT\footnote{ https://huggingface.co/l3cube-pune/hindi-sentence-bert-nli}\cite{joshi2023l3cube}, BengaliSBERT\footnote{ https://huggingface.co/l3cube-pune/bengali-sentence-bert-nli}, MahaSBERT\footnote{https://huggingface.co/l3cube-pune/marathi-sentence-bert-nli}, TeluguSBERT\footnote{ https://huggingface.co/l3cube-pune/telugu-sentence-bert-nli}, TamilSBERT\footnote{ https://huggingface.co/l3cube-pune/tamil-sentence-bert-nli}, GujaratiSBERT\footnote{ https://huggingface.co/l3cube-pune/gujarati-sentence-bert-nli}, KannadaSBERT\footnote{ https://huggingface.co/l3cube-pune/kannada-sentence-bert-nli}, OdiaSBERT\footnote{ https://huggingface.co/l3cube-pune/odia-sentence-bert-nli}, MalayalamSBERT\footnote{ https://huggingface.co/l3cube-pune/malayalam-sentence-bert-nli}, and PunjabiSBERT\footnote{ https://huggingface.co/l3cube-pune/punjabi-sentence-bert-nli}.

\subsection[L3Cube IndicSBERT]{L3Cube Indic SBERT\footnote{https://huggingface.co/l3cube-pune/indic-sentence-bert-nli} (Multilingual)}

IndicSBERT is the first multilingual SBERT model trained specifically for Indic languages. Sentence-BERT (SBERT) \cite{reimers2019sentence} is a modified version of the BERT \cite{devlin2018bert} architecture designed to generate sentence representations for the improved semantic similarity between sentences. The SBERT uses a Siamese network \cite{koch2015siamese} and is trained using specific datasets like STS, resulting in representations specifically geared for semantic similarity.

\subsection[AI4Bharat IndicBERT]{AI4Bharat indicBERT\footnote{https://huggingface.co/ai4bharat/indic-bert}} IndicBERT is a multi-lingual AlBERT model provided by AI4Bharat exclusively
pre-trained in 12 Indian languages. It is pre-trained on AI4Bharat IndicNLP
Corpora of around 9 billion tokens.

\section{\textbf{Evaluation} }
We evaluated each Indic dataset using the monolingual BERT model, multilingual BERT and SBERT models provided by L3Cube, and the indicBERT model provided by AI4Bharat. The results of evaluating these models on the curated datasets are shown in Table 4.

\begin{table}[H] 
 \centering
  \resizebox{82mm}{!}{%
\begin{tabular}{|p{2.0cm}|p{3.0cm}|p{1.5cm}|p{1.5cm}|p{1.5cm}|}
\hline
\textbf{Language} & \textbf{Model} & \textbf{SHC} & \textbf{LDC} & \textbf{LPC}\\

\hline
& HindiBERT& \textbf{86.600} & \textbf{91.681} & \textbf{88.097}\\
    \multirow{4}{*}\centering{Hindi}  & HindiSBERT& 86.518 & 91.534 & 87.810 \\
& IndicSBERT& 85.870 & 91.604 & 86.922 \\
& IndicBERT& 83.910 & 86.809 & 84.424 \\
\hline

& BengaliBERT & 82.549& \textbf{95.293}& \textbf{85.195}
\\
    \multirow{4}{*}\centering{Bengali}  & BengaliSBERT & \textbf{82.749}& 94.141& 84.470
\\
& IndicSBERT&  80.674& 92.457& 85.232
\\
& IndicBERT& 80.007& 90.757& 82.457\\
\hline

& MarathiBERT & \textbf{91.163}& \textbf{94.706} 
& 86.731\\
    \multirow{4}{*}\centering{Marathi}  & MarathiSBERT &  91.017
& 94.349
& \textbf{87.439}\\
& IndicSBERT&  90.510& 93.987
& 87.103\\
& IndicBERT& 89.388& 92.627& 85.222\\
\hline 
 
& TeluguBERT & 89.810 & \textbf{92.765} & \textbf{91.818}\\
    \multirow{4}{*}\centering{Telugu}  & TeluguSBERT & \textbf{90.416} & 92.651 & 91.098 \\
& IndicSBERT& 87.916 & 92.348 & 91.515 \\
& IndicBERT& 88.371 & 92.943 & 91.811 \\
\hline

& TamilBERT & \textbf{81.785} & 84.521 & \textbf{81.300}\\
    \multirow{4}{*}\centering{Tamil}  & TamilSBERT & 81.720 &\textbf{86.122} & 79.573 \\
& IndicSBERT& 81.209 & 84.227 & 79.703\\
& IndicBERT& 81.275 & 83.226 & 80.571 \\
\hline& GujaratiBERT &\textbf{ 89.898} & \textbf{95.278} & \textbf{90.640}\\
    \multirow{4}{*}\centering{Gujarati}  & GujaratiSBERT & 89.808 & 95.060 & 90.091 \\
& IndicSBERT& 88.613 & 95.277 & 89.573 \\
& IndicBERT& 87.909 & 90.854 & 89.050\\
\hline

& KannadaBERT & \textbf{91.410} & 94.706 & \textbf{87.704}\\
    \multirow{4}{*}\centering{Kannada}  & KannadaSBERT & 89.340 & 94.539 & 87.345 \\
& IndicSBERT& 90.255 & \textbf{94.833} & 87.835 \\
& IndicBERT& 87.965 & 91.857 & 87.358 \\
\hline

& OdiaBERT & 83.399 & 92.940 & \textbf{84.557}\\
    \multirow{4}{*}\centering{Odia}  & OdiaSBERT & \textbf{84.137} & \textbf{93.424} & 83.918 \\
& IndicSBERT& 82.065 & 93.054 & 84.441 \\
& IndicBERT& 82.661 & 86.962 & 82.322 \\
\hline

& MalayalamBERT &\textbf{80.440}& \textbf{88.573} & \textbf{81.705}\\
    \multirow{4}{*}\centering{Malayalam}  & MalayalamSBERT & 80.171& 88.201& 81.171\\
& IndicSBERT& 77.780& 88.029& 79.672\\
& IndicBERT& 75.114& 83.098& 76.457\\
\hline

& PunjabiBERT & 85.456 & 90.182 & 84.163\\
    \multirow{4}{*}\centering{Punjabi}  & PunjabiSBERT & 90.363 & \textbf{94.866} & 90.333 \\
& IndicSBERT& 89.490 & 94.224 & 88.446 \\
& IndicBERT& \textbf{91.725}& 94.781 & \textbf{90.358}\\
\hline

& Bert-base uncased & 76.634 & \textbf{92.177} & \textbf{78.392}\\
    \multirow{3}{*}\centering{English}  & IndicSBERT & 75.237 & 91.070 & 77.142 \\
& IndicBERT& \textbf{76.788}& 90.813 & 77.913\\
\hline

\end{tabular}
}

\caption{\label{citation-guide}
 Accuracies for all the models trained on SHC, LDC, and LPC datasets in percentage (\%) 
}
\end{table}

\begin{table}[H] 
 \centering
\resizebox{83mm}{22mm}{%
\begin{tabular}{|p{2.0cm}|p{3.0cm}|p{1.5cm}|p{1.5cm}|p{1.5cm}|}
\hline
\textbf{Language} & \textbf{Model} & \textbf{SHC} & \textbf{LDC} & \textbf{LPC}\\

\hline

    \multirow{4}{*}\centering{Hindi}  
& HindiSBERT& 86.209 & 88.451 & 88.279\\

\hline

    \multirow{4}{*}\centering{Bengali}  & BengaliSBERT &  82.615& 93.126 & 89.289
\\

\hline

    \multirow{4}{*}\centering{Marathi}  & MarathiSBERT &  91.017
& 94.349 & 87.439\\

\hline

    \multirow{4}{*}\centering{Telugu}  & TeluguSBERT & 88.674 & 92.310 & 91.174 \\

\hline

    \multirow{4}{*}\centering{Tamil}  & TamilSBERT & 81.754 & 84.902 & 80.993 \\

\hline

    \multirow{4}{*}\centering{Gujarati}  & GujaratiSBERT & 88.792 & 92.378& 90.609 \\

\hline

    \multirow{4}{*}\centering{Kannada}  & KannadaSBERT & 89.797 & 94.212& 88.260 \\

\hline

    \multirow{4}{*}\centering{Odia}  & OdiaSBERT & 84.137 & 93.424& 83.918 \\

\hline

    \multirow{4}{*}\centering{Malayalam}  & MalayalamSBERT & 79.114& 87.743 & 80.428\\

\hline

    \multirow{4}{*}\centering{Punjabi}  & PunjabiSBERT & 94.744& 93.614 & 93.782 \\

\hline

\multirow{4}{*}\centering{English}  & Bert-base & 74.28 & 94.87 & 75.37 \\

\hline

\end{tabular}
}

\caption{\label{citation-guide}
 Accuracies for monolingual SBERT model trained on mixed dataset (SHC + LPC + LDC) in percentage (\%) 
}
\end{table}

\section{\textbf{Results} }
For evaluating the models on these datasets, we use accuracy as our main evaluation metric to understand the performance of our models. Table 4 presents the accuracies obtained by fine-tuning our models on the datasets. The confusion matrices for the Kannada dataset trained and tested on SHC, LDC, and LPC respectively are illustrated in Figures 4, 5, and 6.  \newline \newline Key observations are outlined as follows: 
\begin{itemize}
    \item In most cases, the L3Cube Monolingual model tends to exhibit superior performance in terms of accuracy, as demonstrated by the metrics presented in the table across all corpora.  
    \item Among the Short Headlines Classification (SHC), Long Document Classification (LDC), and Long Paragraph Classification (LPC) datasets, LDC demonstrated the most impressive results after fine-tuning text classification. This aligns with expectations, given that long document data inherently contains more information compared to shorter-length datasets. 
    \item On the other hand, SHC reported comparatively lower accuracy scores across all three document types. This may be attributed to the fact that news headlines can sometimes encompass more generalized information, potentially leading to some degree of confusion for the models.
    \end{itemize}

\begin{figure}[H]
    \centering
    \includegraphics[width=0.75\linewidth, height=5.5cm]{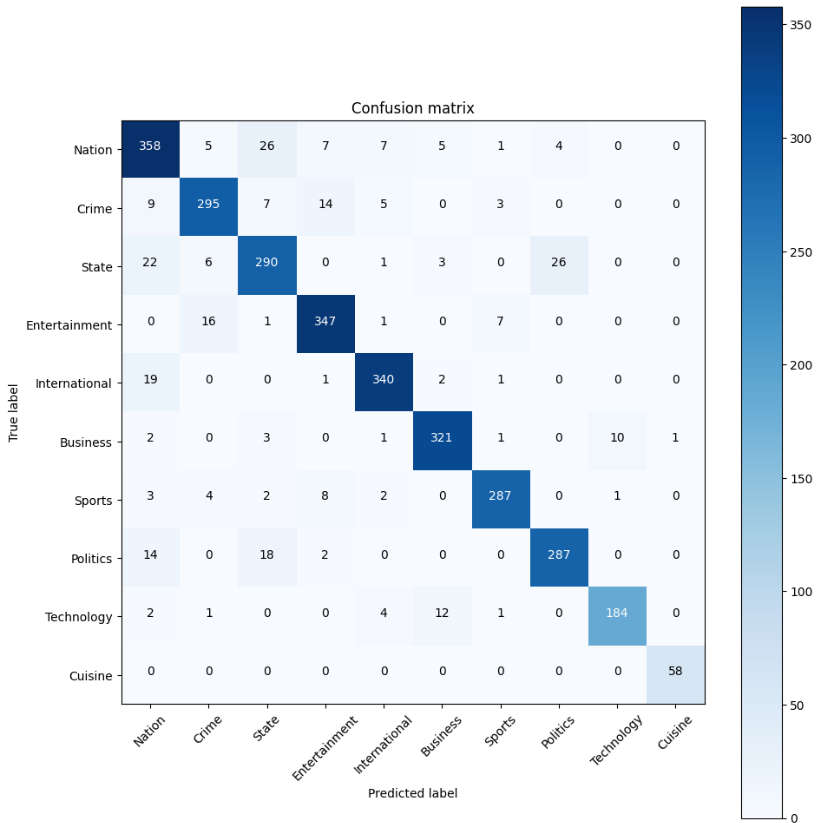}
    \caption{Confusion matrix for the Kannada SHC dataset}
    \label{fig:enter-label}
\end{figure}

\begin{figure}[H]
    \centering
    \includegraphics[width=0.75\linewidth, height=5.5cm]{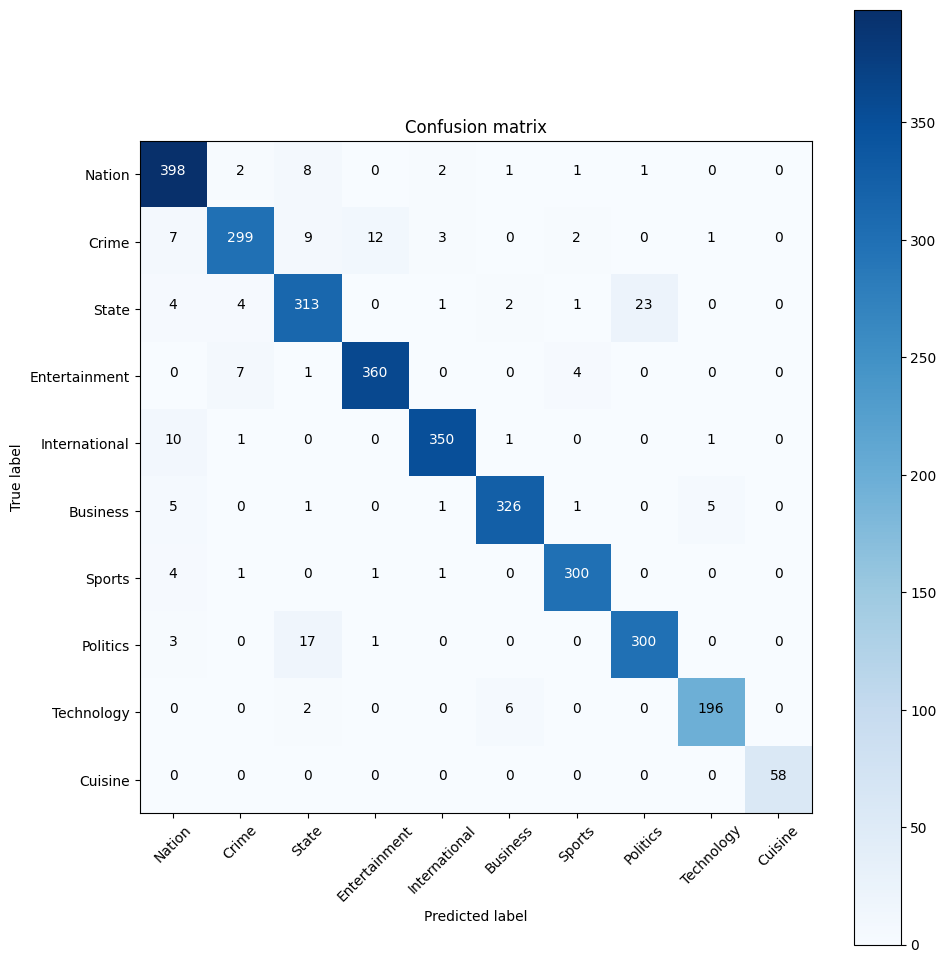}
    \caption{Confusion matrix for the Kannada LDC dataset}
    \label{fig:enter-label}
\end{figure}

\begin{figure}[H]
    \centering
    \includegraphics[width=0.75\linewidth, height=5.5cm]{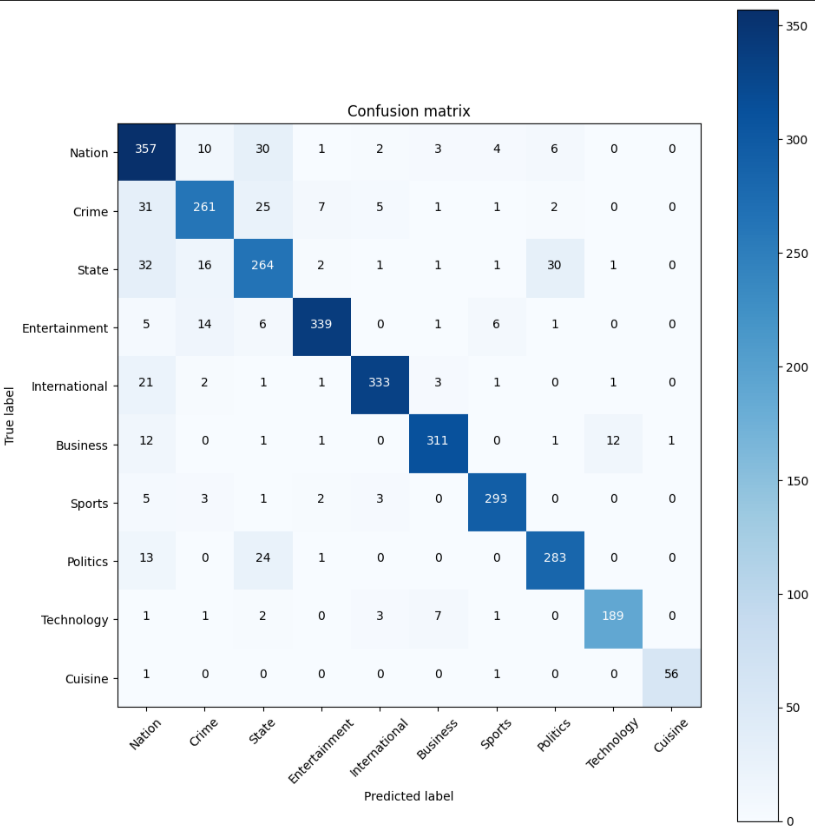}
    \caption{Confusion matrix for the Kannada LPC dataset}
    \label{fig:enter-label}
\end{figure}

Table 5 presents the accuracies obtained by training the L3Cube Monolingual Sentence BERT model on a mix of all the 3 datasets (SHC + LPC + LDC) and then evaluating them on individual datasets. A cross-analysis was conducted by evaluating the performance of L3Cube Monolingual Indic Sentence BERT model on individual test datasets. Upon fine-tuning the Monolingual SBERT model on a mixed dataset, notable results were observed, with the Long Document Classification (LDC) task exhibiting the most impressive performance.

\section{\textbf{Future Work} }

There is potential for expanding the dataset's labels to achieve broader category coverage or facilitate further category expansion. The dataset creation process has been streamlined through automation, eliminating the need for manual typing. Looking forward, there is a prospect of implementing the curation of a manually typed or verified dataset, ensuring an even higher level of accuracy and reliability for future applications. 

Currently, we have achieved proficiency in Long Document Classification (LDC) for a single language, and the performance in Long Paragraph Classification (LPC) is satisfactory. However, we acknowledge that the performance in Short Headlines Classification (SHC) falls short of optimization. To address this, we have implemented a unified model selection, specifically opting for the monolingual SBERT based on its significant performance on individual datasets for each language.

This chosen model exhibits competence in handling mixed datasets, comprising of SHC, LPC, and LDC, and its performance has been thoroughly examined. It's important to note that our current focus remains on a single language. Looking ahead, we recognize the potential for future developments, particularly in the realm of cross-dataset analysis. 

\section{\textbf{Conclusion} }
In this research paper, we introduce L3Cube-IndicNews, a comprehensive collection of three labeled datasets, encompassing over 3 lakh records in ten different Indic languages. These datasets are designed for text classification tasks in the context of Indian languages. Within this paper, we provide an in-depth overview of the creation process, which involves the use of 10 to 12 distinct categorical labels to curate these supervised datasets. To assess the effectiveness of these datasets, we conducted fine-tuning on BERT-based models, serving as a valuable benchmark for future research and development.

Our experiments involved four key models: L3Cube Monolingual BERT, L3Cube Monolingual SBERT, L3Cube-IndicSBERT (multilingual),  and AI4BHARAT IndicBERT. Notably, our findings indicate that the BengaliBERT Model achieved the highest accuracy when applied to the LDC dataset. We believe that the availability of our datasets will contribute significantly to the enhancement of NLP support for the Indic languages, promoting its growth and development in this field.

\section{\textbf{Acknowledgements} }
This work was carried out under the mentorship of L3Cube, Pune. We would like to express our gratitude towards our mentor, for his continuous support and encouragement. This work is a part of the L3Cube-IndicNLP Project.

\bibliography{main}
\newpage
\section{\textbf{Appendix} }

\begin{table} [H]
\centering
\begin{tabular}{|c|c|c|}
\hline
\textbf{Language} & \textbf{Model Name} & \textbf{Url}\\
\hline

Hindi&HindiSBERT&\href{https://huggingface.co/l3cube-pune/hindi-topic-all-doc}{hindi-topic-all-doc}\\
\hline

 Bengali&BengaliSBERT&\href{https://huggingface.co/l3cube-pune/bengali-topic-all-doc}{bengali-topic-all-doc}  \\
\hline

Marathi&MarathiSBERT&\href{https://huggingface.co/l3cube-pune/marathi-topic-all-doc-v2}{marathi-topic-all-doc-v2}\\
\hline

 Telugu&TeluguSBERT&\href{https://huggingface.co/l3cube-pune/telugu-topic-all-doc}{telugu-topic-all-doc}\\
\hline

Tamil& TamilSBERT&\href{https://huggingface.co/l3cube-pune/tamil-topic-all-doc}{tamil-topic-all-doc} \\
\hline

 Gujarati&GujaratiSBERT&\href{https://huggingface.co/l3cube-pune/gujarati-topic-all-doc}{gujarati-topic-all-doc}\\
\hline

 Kannada& KannadaSBERT&\href{https://huggingface.co/l3cube-pune/kannada-topic-all-doc}{kannada-topic-all-doc}\\
 \hline
 
Odia&OdiaSBERT&\href{https://huggingface.co/l3cube-pune/odia-topic-all-doc}{odia-topic-all-doc} \\
\hline

 Malayalam&MalayalamSBERT& \href{https://huggingface.co/l3cube-pune/malayalam-topic-all-doc}{malayalam-topic-all-doc}\\
\hline

 Punjabi&PunjabiSBERT &\href{https://huggingface.co/l3cube-pune/punjabi-topic-all-doc}{punjabi-topic-all-doc}\\
\hline
English&BERT-Based-Uncased &\href{https://huggingface.co/l3cube-pune/english-topic-all-doc}{english-topic-all-doc}\\

\hline

\end{tabular}
\caption{\label{citation-guide}
 Link to the models on Hugging Face 
}
\end{table}

\end{document}